# Pretrained battery transformer (PBT): a foundation model for battery life prediction

Ruifeng Tan,[ab] Weixiang Hong,[a] Jia Li,*[c] Jiaqiang Huang*[ade] and Tong-Yi Zhang*[fg]

Early prediction of battery cycle life is essential for improving battery design, manufacturing and deployment. However, despite encouraging progress with machine learning, battery life prediction remains constrained by scarce data and pronounced heterogeneity across battery chemistries, specifications, formation protocols and operating conditions. Although transfer learning has been widely explored to alleviate these challenges, its effectiveness is limited by the absence of a foundation model that can integrate heterogeneous battery life data and provide broadly useful knowledge for target-scenario specialization. Here we introduce the pretrained battery transformer (PBT), an integrated foundation model comprising a general PBT and specialized PBT models for individual target scenarios. At its core, battery-knowledge-encoded mixture-of-experts layers enable the general PBT to consolidate shared cycling-pattern–lifetime relationships from 13 heterogeneous lithium-ion battery datasets while preserving specialization across distinct aging regimes. The resulting general PBT provides a shared knowledge and parameter foundation from which specialized PBT models are constructed using limited labelled data to capture target-specific degradation behavior. Across 15 downstream datasets covering 977 batteries and 532 aging conditions from lithium-ion, sodium-ion and zinc-ion batteries, the specialized PBT models achieve state-of-the-art performance, outperforming the strongest comparator by 24.8% on average and by up to 73.9%. This study establishes, to our knowledge, the first foundation model for battery life prediction and points towards a shift from isolated, scenario-specific modelling to a reusable knowledge foundation for data-efficient specialization, with broader implications for sustainable-energy prediction problems constrained by scarce and heterogeneous data.

## Broader context

The rapid expansion of electric vehicles and grid-scale storage, together with the diversification of battery designs and operating conditions, is making lifetime evaluation a central bottleneck in energy-storage innovation. Each new material, format, or operating protocol can require months to years of battery cycling tests before lifetime can be reliably determined, while the resulting data are often sparse, fragmented, and difficult to transfer across battery systems. Although transfer learning has been explored to mitigate this challenge, existing approaches remain limited by the lack of a general pretrained model capable of consolidating heterogeneous battery aging data. Here, we introduce the pretrained battery transformer (PBT), an integrated foundation model that incorporates battery knowledge and comprises a general PBT and specialized PBT models. The general PBT consolidates cycling-pattern–lifetime mappings from aging data into a reusable knowledge and parameter foundation. Specialized PBT models build on the general model using limited data to capture scenario-specific degradation behavior. This framework provides a practical route to data-efficient lifetime prediction across battery design, manufacturing and deployment, and highlights the need for energy foundation models that combine data scale with domain knowledge to make scarce, costly and heterogeneous experimental data more reusable.

[a] *Guangzhou Municipal Key Laboratory of Materials Informatics and Sustainable Energy and Environment Thrust, The Hong Kong University of Science and Technology (Guangzhou), Nansha, Guangzhou, 511400, Guangdong, P. R. China. E-mail: seejhuang@hkust-gz.edu.cn*
[b] *Department of Computer Science & Engineering, The Hong Kong University of Science and Technology, Clear Water Bay, Kowloon, 999077, Hong Kong, P. R. China*
[c] *Guangzhou Municipal Key Laboratory of Materials Informatics and Data Science and Analytics Thrust, The Hong Kong University of Science and Technology (Guangzhou), Nansha, Guangzhou, 511400, Guangdong, P. R. China. E-mail: jialee@ust.hk*
[d] *Academy of Interdisciplinary Studies, The Hong Kong University of Science and Technology, Clear Water Bay, Kowloon, 999077, Hong Kong, P. R. China*
[e] *Guangzhou HKUST Fok Ying Tung Research Institute, Nansha District, Guangzhou, 511458, Guangdong, P. R. China*
[f] *Material Genome Institute, Shanghai University, 333 Nanchen Road, Shanghai, 200444, P. R. China*
[g] *Guangzhou Municipal Key Laboratory of Materials Informatics, Advanced Materials Thrust, and Sustainable Energy and Environment Thrust, The Hong Kong University of Science and Technology (Guangzhou), Nansha, Guangzhou, 511400, Guangdong, P. R. China. E-mail: mezhangt@hkust-gz.edu.cn*

## Introduction

Rechargeable batteries are central to the low-carbon energy transition, underpinning electric vehicles, grid storage and portable electronic devices.[1–5] In 2024, the global shipment of LIBs alone exceeded 1545 GWh, with total production projected to reach 4700 GWh by 2030.[6–8] In addition to LIBs, alternatives such as sodium-ion, zinc-ion and full-solid-state batteries are attracting attention because of their potential advantages in terms of cost and safety.[9–11] Nevertheless, all rechargeable batteries inevitably degrade because of the effects of intrinsic electrochemical processes,[12–14] leading to capacity fading, stability loss and safety concerns in practical applications.[15–17] Thus, aged batteries are retired from high-demand applications such as electric vehicles when they reach the end of their life cycle.[18–20] Battery lifetime is typically measured based on the number of charge–discharge cycles at which the battery capacity falls to 80% of the nominal capacity.[12,16] Traditionally, determining battery lifetime requires full-life cycling experiments that can take months or even years, creating a major bottleneck for battery development and evaluation.[13] Accurate early battery life prediction is therefore essential for optimizing battery design, production and utilization.

Data-driven methods have yielded impressive results in early battery life prediction.[12,13,16,21–28] Early approaches typically rely on domain knowledge to construct features to capture lifetime-relevant signatures associated with interfacial processes,[29] manufacturing inconsistencies[24] and other latent degradation processes,[30,31] all of which impact battery degradation behavior. These features were derived from voltage–current profiles,[16,23,26,28,32] physics-based model outputs,[33] or complementary measurements such as surface temperature[25] and electrochemical impedance spectra.[34] The derived features were subsequently used as inputs to machine-learning models for lifetime prediction. For example, differential-voltage features characterized graphite-anode manufacturing heterogeneity associated with cell-to-cell differences in cycle life and served as inputs to a ridge regression model for battery lifetime prediction.[24] Although effective for some datasets, these feature-engineering approaches often rely on handcrafted features, additional measurements or prior knowledge that may be unavailable or inapplicable across battery chemistries and operating conditions,[12,21,22,24] limiting their broader applicability. Neural networks provide a more general alternative by learning feature representations directly from early-cycle measurements, with their greater expressive capacity often enabling them to outperform approaches relying on handcrafted features.[12,13,21,22,35,36] Among the neural networks, CyclePatch models were identified as leading models among 18 competitive neural networks by capturing inherently recurring patterns across cycles in cycling data.[13] Despite these advances, battery lifetime prediction remains hindered by two fundamental challenges: the scarcity of lifetime data and the pronounced heterogeneity introduced by variations in cathode and anode materials, electrolytes, manufacturing processes, formation protocols and operating conditions.[12,13] In particular, data heterogeneity leads to variations in voltage–current profiles, producing a distribution shift between training and testing data—a phenomenon known in machine learning as domain shift—that often degrades model performance.[5,37] Consequently, limited data can prevent models trained from scratch from learning reliable representations, whereas models trained on one dataset may still perform poorly on out-of-distribution battery datasets. Therefore, both data scarcity and heterogeneity must be considered in the design of practical battery lifetime prediction models.

A promising route to addressing the coupled challenges of data scarcity and heterogeneity is to exploit cycling-pattern–lifetime relationships that can be reused across battery datasets. Such cross-dataset knowledge reuse allows observations from related battery systems and aging conditions to strengthen lifetime prediction in data-limited scenarios, thereby increasing the effective information available for model development. Transfer learning, typically implemented through fine-tuning[21,38] or domain adaptation,[39] has been explored as a promising means of realizing such reuse by incorporating knowledge learned from source datasets into prediction models for target datasets. For instance, a recent study introduced BatLiNet,[12] which uses LFP/graphite cells as reference batteries to construct inter-cell differences and incorporates these differences into learned representations for lifetime prediction across aging conditions. Previous theoretical studies[40,41] suggest that target-domain performance is jointly constrained by the extent to which the source-domain distributions cover degradation patterns relevant to the target domain and the predictive performance of the source model. In this vein, despite encouraging results, existing approaches to learning and exploiting reusable cycling-pattern–lifetime relationships remain limited because their source models were not designed to consolidate diverse battery lifetime data into a general model. These constraints underscore a fundamental need for a general pretrained model that can learn broadly applicable cycling-pattern–lifetime relationships from heterogeneous datasets and provide a unified parameter foundation for target-scenario specialization.

Foundation models (FMs) have emerged as pretrained models that encode general knowledge and can serve as a common basis for diverse downstream applications.[42,43] These FMs exhibit broad utility and constitute indispensable components of many state-of-the-art natural language processing (NLP) and computer vision (CV) systems.[44–48] Nevertheless, their success depends on architectures tailored to the structure of domain-specific data. For example, DeepSeek[49,50] and GPTs[51,52] employ transformer variants optimized for text, whereas CV FMs[53–56] use convolutional neural networks or vision transformers. However, effective architectures for encoding general battery lifetime knowledge are currently lacking. Experimental battery lifetime data are highly heterogeneous because battery degradation is shaped by diverse specifications, formation histories and operating conditions, spanning electrode and electrolyte materials,[57–59] chemistries,[13] cell formats,[13] formation protocols,[26,60] depths of discharge,[57–59,61] charge and discharge protocols,[21,57,58,62–65] stress factors[61] and operating temperatures.[59,61] At the same time, these datasets are limited in scale: even large single-scenario battery aging datasets typically contain only tens of batteries, and aggregated public lifetime datasets remain orders of magnitude smaller than the text and image corpora used to train general-purpose NLP and CV FMs. Building a battery FM for lifetime prediction therefore requires architectures that can organize heterogeneous lifetime information into a reusable knowledge foundation under data scarcity. Moreover, the parameterization learned from heterogeneous data must remain amenable to specialization using only limited target observations, thereby enabling the model to capture scenario-specific degradation behavior.

Here we present, to our knowledge, the first foundation model for battery cycle life prediction that operates directly on commonly available voltage–current data: the pretrained battery transformer (PBT). At its core is BatteryMoE, a mixture-of-experts layer that incorporates battery knowledge through soft and hard encoders to guide learning under data scarcity, organize shared cycling-pattern–lifetime relationships across heterogeneous datasets and preserve expert specialization for distinct aging regimes. By pretraining this architecture on 13 lithium-ion battery datasets, we construct a general PBT that encodes broadly applicable battery lifetime knowledge. As a whole, PBT comprises this general PBT and a set of specialized PBTs constructed from the general model's learned parameterization using limited data from individual target datasets. Each specialized PBT can thereby capture target-specific degradation behavior while building on the knowledge acquired through heterogeneous pretraining. Across 15 downstream datasets covering 977 batteries and 532 aging conditions, the specialized PBTs achieve state-of-the-art performance for lithium-ion datasets within the pretraining-covered specifications as well as sodium-ion, zinc-ion and large-format industrial lithium-ion datasets beyond these specifications. Notably, competitive neural baselines pretrained on the same data and specialized using the same procedures do not achieve comparably consistent performance improvements, highlighting the importance of BatteryMoE in consolidating heterogeneous battery lifetime data and supporting effective target-specific modelling. Finally, ablation and data-scaling analyses clarify the respective contributions of aging-condition information, knowledge-guided expert routing and pretraining-data coverage to PBT performance. Together, these results establish PBT as a practical battery foundation model that advances battery lifetime prediction beyond isolated, scenario-specific modelling by transforming heterogeneous aging data into a reusable knowledge and parameter foundation for data-efficient specialization across various scenarios.

## Datasets

To provide a sufficiently broad empirical foundation for learning shared cycling-pattern–lifetime relationships across heterogeneous battery systems, we compiled the largest and most diverse battery life database to date, consisting of 16 publicly available datasets[13,15,16,21,26,58–62,64,66–69] (see SI Note S1 for the data pre-processing details and SI Table S1 for the statistics of each dataset), including various formation protocols,[26,60] depths of discharge,[57–59,61] charge protocols,[15,16,57–59,64–66,70] discharge protocols,[21,57,58,62–65] stress factors,[61] materials,[62–64] operating temperatures,[59,61] chemistries of LIBs and other factors.[13] As shown in Fig. 1a, in comparison with the previous largest database,[12] our database includes 2.61 times as many batteries, 4 times as many physical formats, 8.71 times as many chemical systems, 1.8 times as many operating temperatures, 2.2 times as many charge–discharge protocols, and 2.89 times as many aging conditions. For clarity, throughout this work, an aging condition is defined by a unique combination of 10 aging factors (battery physical format, anode, cathode, electrolyte, nominal capacity, manufacturer, formation protocol, charge protocol, discharge protocol, and operating temperature), such that any difference in one or more factors defines a distinct aging condition.

Despite its unprecedented scale, our database still encodes data scarcity and heterogeneity challenges, as it contains information for only 977 batteries spanning 532 aging conditions, of which 418 are represented by single cells (Fig. 1b). The database therefore provides a rigorous testbed for evaluating whether a foundation model for battery lifetime prediction can be established by consolidating scarce and heterogeneous lifetime data into a general PBT and using the resulting knowledge and parameter foundation to construct specialized PBT models for diverse aging scenarios.

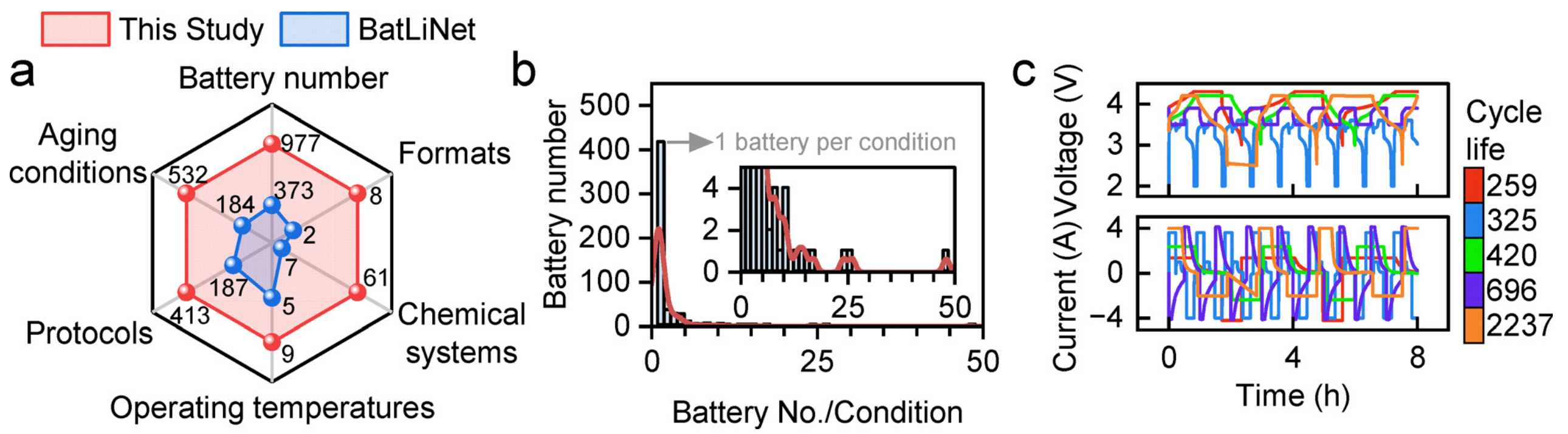

Fig. 1 Data used in this study and challenges arising from data scarcity and heterogeneity. (a) Comparison of data used in this study and those used in the second-largest battery cycle life prediction study (with BatLiNet[12]). Notably, both datasets were processed using the same pipeline (see SI Note S1) and the filtered batteries were excluded from the statistical analysis.[13] In this work, we consider the battery physical format, anode, cathode, electrolyte, nominal capacity, manufacturer, formation protocol, charge protocol, discharge protocol, and operating temperature as aging factors. Any difference in one or more factors constitutes a distinct aging condition. (b) Distribution of the number of batteries across 532 aging conditions, showing that more than 400 aging conditions are represented by a single battery. (c) Cycling patterns of five batteries under five aging conditions (see SI Note S2) during the first eight hours of use.

**Overview of the pretrained battery transformer (PBT)**

Our goal is to develop a unified foundation model (FM) for early cycle-life prediction across a broad range of practically important settings,[12,22,25,36] including battery screening, manufacturing evaluation, protocol design and deployment under diverse operating conditions. We formulate PBT as an integrated framework comprising a general PBT and specialized PBT models. The general PBT is pretrained on batteries with known lifetime labels collected under diverse aging conditions, providing a shared knowledge and parameter foundation. Specialized PBT models are constructed from this foundation using limited labelled data from individual target scenarios. Both forms predict the cycle life of a new battery from voltage–current data collected during its first $N$ charge–discharge cycles, where $N$ can be any positive integer up to 100. We set this upper limit because previous studies[12,16] have shown that this range is both practically relevant and experimentally feasible for early-cycle lifetime prediction. Cycle life is generally defined as the number of cycles at which discharge capacity declines to 80% of nominal capacity. For the industrial Li-ion battery dataset considered in this study, we instead use a 90% threshold because most batteries in this dataset were not degraded to 80% of nominal capacity.

Learning such a shared foundation is challenging because battery lifetime data are scarce and highly heterogeneous. Variations in materials, manufacturing processes, formation protocols and operating conditions can produce substantially different early-cycle behaviors and lifetime outcomes (Fig. 1c). A practical battery FM must therefore identify cycling-pattern–lifetime relationships that can be reused across datasets while retaining the capability to represent degradation behavior specific to individual aging regimes.

To meet this requirement, we develop BatteryMoE as the core module of PBT. BatteryMoE incorporates battery knowledge into aging-condition-dependent expert routing, enabling PBT to organize shared cycling-pattern–lifetime relationships across heterogeneous datasets while preserving expert specialization for distinct aging regimes. Rather than representing all batteries through a single homogeneous mapping, the gate network applies aging-condition-dependent routing to direct cycling representations to appropriate expert networks and combines their outputs using learned weights. Batteries with related attributes can therefore learn through overlapping expert pathways, whereas distinct degradation patterns remain represented by specialized pathways. This structure enables efficient learning from scarce and heterogeneous data and provides the general PBT with a suitable knowledge foundation for constructing specialized PBT models.

BatteryMoE comprises two main components: a battery-knowledge-encoded gate network and a set of battery-knowledge-informed expert networks (Fig. 2a). It builds on the mixture-of-experts (MoE) paradigm, in which a gate network routes inputs to specialized neural networks and which underlies many modern foundation models.[71] Conventional MoE architectures learn expert assignments solely from data, which can be ineffective for battery lifetime prediction because of the limited data volume and complex voltage–current dynamics. BatteryMoE therefore introduces battery knowledge through two complementary encoders. The soft encoder converts aging-condition information into representations that guide the initial expert weights. The hard encoder subsequently imposes physically meaningful constraints on these weights by retaining only the experts that are consistent with known battery attributes. Their combination retains flexible, data-driven weighting within a battery-informed expert structure, allowing PBT to learn shared relationships without erasing specialization across aging regimes.

The soft encoder (pink region in Fig. 2a) supplies the gate network with embeddings derived from aging-condition descriptions rather than raw voltage–current inputs. Ten key aging factors that shape the observed cycling-data patterns are organized into three groups: cell specifications, comprising the cathode, anode, electrolyte, nominal capacity, physical format and manufacturer; formation conditions, represented by the formation protocol; and operating conditions, comprising the charge protocol, discharge protocol and temperature. PBT represents these factors using a unified text-based template (SI Fig. S1). This representation accommodates previously unseen aging-factor values without predefined vocabularies or manually designed mapping rules, allowing the model to incorporate diverse aging-factor values that can be described in natural language. Missing information is represented consistently by marking the corresponding factor as "unknown", without changing the input structure or model architecture. The completed template is processed by an LLM-based aging network, which converts the textual description into a vector representation. These representations exhibit meaningful clustering across batteries (SI Fig. S2), indicating that they capture certain similarities and distinctions among aging conditions that are useful for expert routing. A linear layer then maps each representation to the initial expert weights. Additional aging factors, such as stacking stress[72] and coating processes,[73] could be incorporated in the same manner, although they are not considered here because they are absent from the collected datasets.

The hard encoder imposes physically grounded routing constraints on the initial expert weights produced by the soft encoder, thereby reducing the routing space that must be learned from limited data. We consider four aging factors that can be categorized using well-established battery knowledge: cathode, anode, physical format and operating temperature. Other aging factors lack broadly accepted categorical definitions and are therefore not explicitly hard-coded. For the categorical factors—cathode, anode and physical format—BatteryMoE constructs experts for individual categories, such as LFP and $Li(Ni_xCo_yAl_{1-x-y})O_2$ (NCA) cathodes, and retains only those compatible with the input battery, because batteries sharing these attributes often exhibit related degradation behaviors.[59,74] Temperature is treated as a continuous variable: experts are defined at discrete temperatures, and only those falling within ±5 °C of the input temperature are retained, reflecting the similarity of degradation mechanisms over nearby temperature ranges.[75–77] For example, for an 18 650 LFP/graphite cell cycled at 25 °C, the LFP, graphite, 18 650-format and 20–30 °C temperature experts are retained.

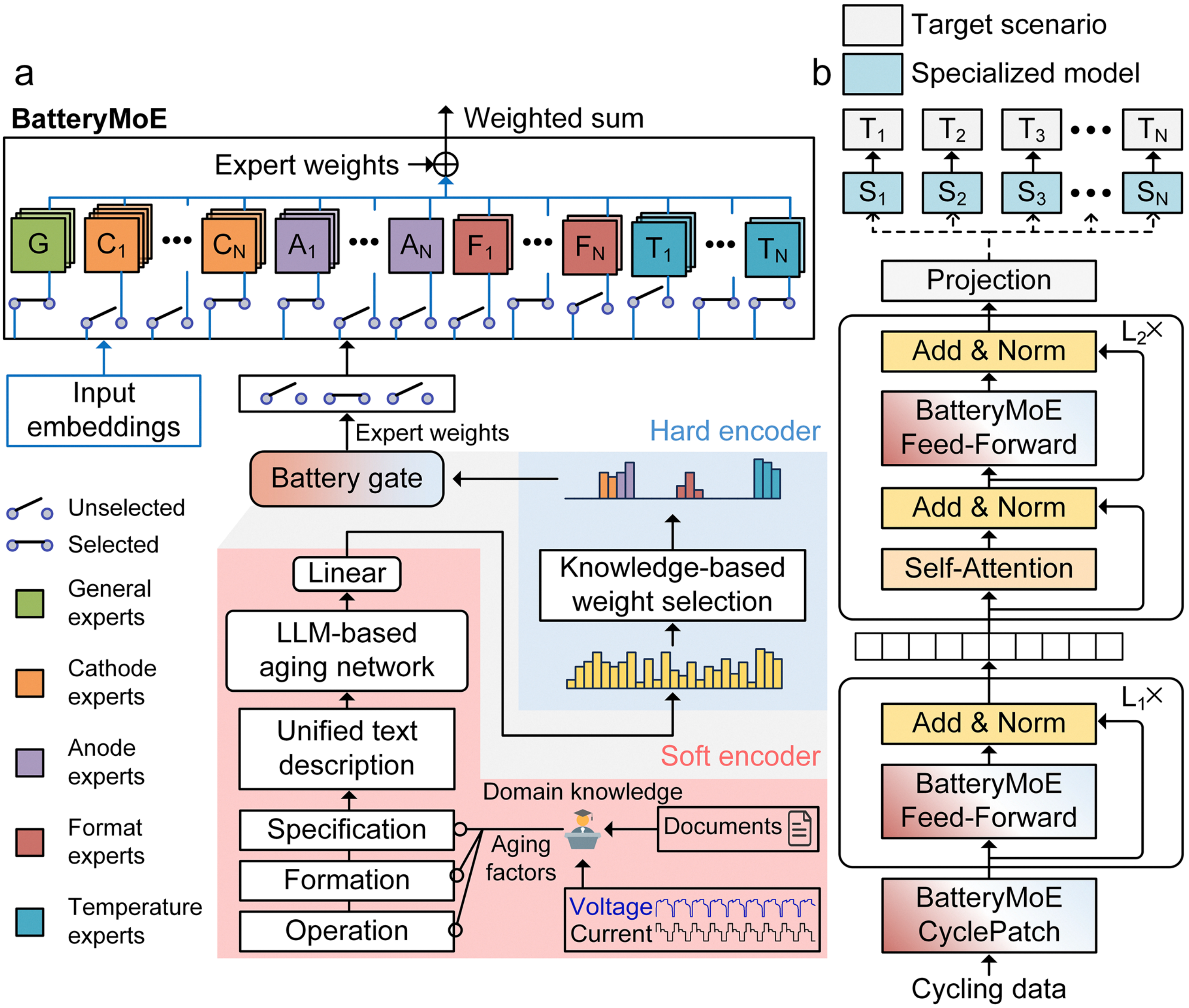

Fig. 2 Overview of BatteryMoE and PBT. (a) BatteryMoE architecture. A battery-knowledge-encoded gate network integrates information from the soft and hard encoders to generate aging-condition-dependent weights, route input embeddings to selected experts and aggregate their outputs. (b) Overview of the PBT framework, comprising a general PBT and a series of specialized PBT models adapted from the general PBT for various target scenarios.

In implementation, the hard encoder retains the initial weights associated with experts that match the input battery. The retained experts operate in parallel to extract lifetime-relevant representations, and their outputs are aggregated using learned weights. Each category can contain multiple experts, providing finer specialization for aging factors that are not explicitly hard-coded, such as formation protocols and operating conditions. This arrangement allows batteries with shared attributes to use overlapping expert pathways while retaining separate modelling capacity for distinct degradation regimes. BatteryMoE therefore organizes heterogeneous cycling-pattern–lifetime relationships through battery-knowledge-guided, aging-condition-dependent routing and structured expert specialization. Because different neural network types can serve as experts, BatteryMoE can also function as a plug-in module compatible with architectures beyond the specific PBT implementation considered here.

As shown in Fig. 2b, BatteryMoE modules constitute the core computational units of PBT. Any module whose name includes "BatteryMoE" denotes a BatteryMoE layer in which aging-condition-dependent routing and expert selection follow the BatteryMoE design. For example, the "BatteryMoE feedforward" module is a BatteryMoE layer whose expert networks adopt a feedforward architecture. Each charge–discharge cycle is first encoded using "BatteryMoE CyclePatch",[13] which converts the cycling data into a cycle token. The resulting tokens are progressively refined through stacked BatteryMoE feedforward layers with residual connections and are then passed to a transformer encoder in which the conventional feedforward layers are replaced by BatteryMoE modules. A projection head maps the encoder output to the predicted battery cycle life. This architecture provides the common backbone for both the general PBT and the specialized PBT models.

Pretraining yields the general PBT, which captures broadly applicable lifetime knowledge. For each target application, its parameterization is adapted using limited target data through transfer learning to construct a specialized PBT (Fig. 2b). PBT thus comprises the general and specialized models, with BatteryMoE organizing shared knowledge while preserving aging-regime specialization.

# Results

### The general PBT consolidates comprehensive battery lifetime knowledge from heterogeneous datasets

Within PBT, the general model provides the shared knowledge and parameter foundation from which specialized PBT models are constructed through transfer learning. Transfer-learning theory indicates that target-domain performance is jointly constrained by the predictive capability of the source model and the extent to which the source-domain distributions cover degradation patterns relevant to the target domain.[40,41] Having assembled a database encompassing degradation patterns across diverse aging conditions, the question here is whether general PBT can consolidate these patterns into an accurate general model. We therefore first evaluate the comprehensiveness of the battery lifetime knowledge encoded by the general PBT across the diverse pretraining datasets, before examining the construction of specialized PBT models for individual target scenarios.

To address this question, we pretrained general PBT on the training sets of 13 LIB datasets drawn from the abovementioned database. These datasets comprised 820 batteries spanning 417 aging conditions, 398 charge–discharge protocols, 8 operating temperatures, 15 chemical systems and 5 physical formats, covering most commercial LIB chemistries (see SI Table S2 for the chemistries covered by each dataset). This broad coverage enables a systematic evaluation of whether general PBT can extract lifetime-relevant knowledge across diverse battery specifications, protocols and degradation regimes. Each dataset was randomly partitioned into training, validation, and test subsets at a ratio of 6 : 2 : 2. Model training was repeated three times using different random seeds. The test subsets were used to evaluate the knowledge acquired by the models during pretraining. Test-set aging conditions were considered as seen if they were present in either the training or validation subset and as unseen if they were absent from both. Note that the UL-PUR dataset included information for only 2 batteries after data preprocessing; therefore, no corresponding test set was established in this work. For all analyses, the mean absolute percentage error (MAPE) was employed as the evaluation metric because battery lifetime varies substantially across datasets (see SI Fig. S3 for the battery life distributions from different datasets), ranging from 102 to 4999 cycles; thus, MAPE avoids bias associated with different lifetime scales. Specifically, MAPE was first computed within each aging condition and then averaged across conditions, avoiding overrepresentation of conditions containing more batteries. We compare the general PBT against two baselines (CPMLP and CPTransformer) that were identified as the leading models among 18 neural networks,[13] including iTransformer,[78] PatchTST,[79] MICN,[80] and DLinear.[81] Both baselines employ CyclePatch[13] to generate cycle-level embeddings but differ in how they model inter-cycle relationships. CPMLP uses multilayer perceptrons to capture interactions across cycles, whereas CPTransformer leverages a transformer encoder. Because all models are trained and evaluated using the same pretraining data, their performance differences isolate whether the BatteryMoE-based architecture enables the general PBT to organize heterogeneous battery lifetime knowledge more effectively than conventional neural architectures.

First, we compare the overall results of the general PBT with those of the leading baselines. As shown in Fig. 3a, general PBT achieves the lowest MAPE, with an overall MAPE of 0.151, surpassing that of the second-best model (CPMLP) by 22.7%. To assess its robustness across data regimes, we further evaluate the performance under 60 seen and 55 unseen aging conditions, which correspond to in-distribution and out-of-distribution scenarios, respectively. The general PBT outperforms the second-best model by 22.6% under seen conditions and 22.8% under unseen conditions, indicating that its superiority arises from consistent improvements for both in-distribution and out-of-distribution data. Moreover, we evaluate general PBT across different numbers of useable cycles to examine how effectively the pretrained models extract useful information from the early operational data under different early prediction settings. General PBT maintains clear advantages under all settings, achieving up to a 30.5% improvement when only information from the first cycle is available (Fig. 3b). Remarkably, the general PBT supplied with only the first cycle still outperforms the second-best baseline supplied with 100 cycles, demonstrating its more effective use of limited early-cycle information.

Next, we evaluate model performance on individual datasets. Because different datasets generally involve different battery specifications, protocols and degradation regimes, this analysis probes the breadth of lifetime knowledge captured by the pretrained models. General PBT achieves the best results for 9 out of 12 datasets, with reductions in MAPE of more than 15% relative to the other methods for 8 datasets (Fig. 3c). For 3 datasets (RWTH,[68] Tongji,[64] and CALCE[66,70]), general PBT does not yield the best performance, but it still performs on par with the best model for 2 of them (RWTH and Tongji).

Collectively, these evaluations show that the general PBT consolidates broader and more consistently useful battery lifetime knowledge than strong neural baselines trained on the same heterogeneous data. Its performance across seen and unseen aging conditions, individual datasets and different numbers of useable cycles establishes the shared knowledge and parameter foundation from which specialized PBT models are constructed in the following sections.

### Building specialized PBT models for target scenarios within pretraining-covered battery specifications

Having established that the general PBT consolidates comprehensive battery lifetime knowledge, we next evaluate the effectiveness of specialized models constructed from this shared knowledge and parameter foundation for individual target datasets. We first focus on 12 target datasets whose battery specifications are represented in the pretraining data. For each dataset, we construct a specialized PBT from the general PBT using the available target-dataset training data. Fine-tuning and adapter tuning[46] are considered as two alternative specialization strategies, with the strategy yielding the best validation performance used for each target dataset (see SI Note S3). These target datasets represent practically important scenarios, including fast-charging protocol optimization (MATR[15,16]), formation protocol analysis (Stanford[60]), cycle life evaluation under stress factors (MICH-EXP[61]), and lifetime prediction across diverse operating conditions (other datasets[21,26,57–59,62,64–68]). For a fair comparison, the same specialization strategies are applied to the pretrained CPMLP and CPTransformer models to construct their specialized variants, CPMLP-T and CPTransformer-T. We additionally compare these specialized models with (i) their pretrained counterparts used without target-dataset specialization and (ii) models trained from scratch using only the corresponding target-dataset training data, denoted by the suffix "-S". These comparisons distinguish the contributions of the pretrained knowledge foundation, target-dataset specialization and target-only training. Specialized PBT is also compared with BatLiNet,[12] a recently introduced transfer learning model for battery lifetime prediction (see SI Note S4 for the implementation details).

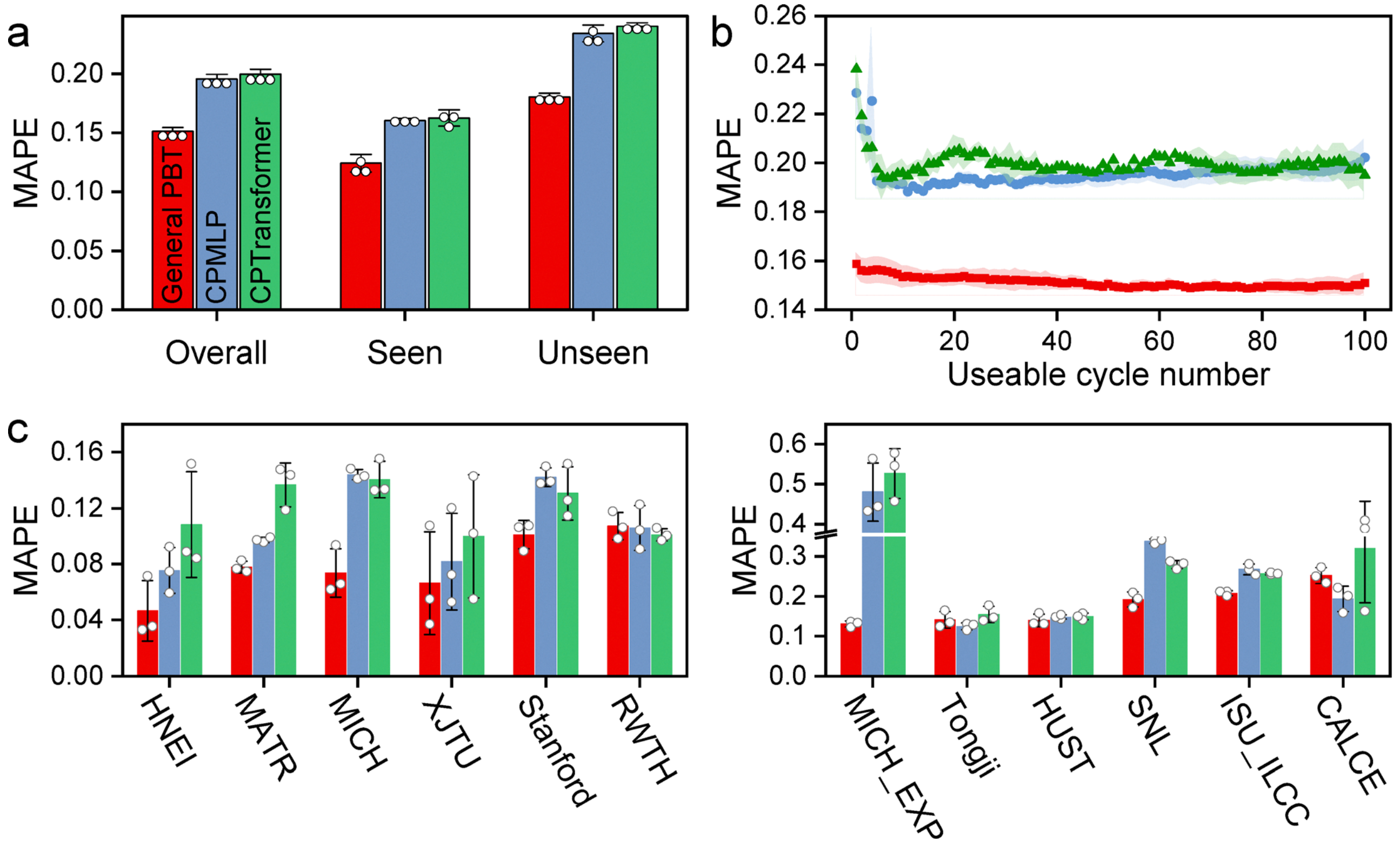

Fig. 3 Evaluation of the comprehensiveness of battery lifetime knowledge encoded by the general PBT. (a) Mean absolute percentage error (MAPE) on the test sets of the 13 LIB datasets used for pretraining. "Overall" denotes the MAPE aggregated across all aging conditions in these test sets. "Seen" and "Unseen" denote results for aging conditions that are present or absent, respectively, from the training and validation subsets of the pretraining data. (b) Model performance as a function of the number of useable cycles. Shaded areas indicate standard deviations across three independent runs. (c) Model performance on the test set of each pretraining dataset. Error bars indicate standard deviations across three independent runs.

Across all 12 target datasets, the specialized PBT achieved the lowest MAPE, outperforming the best-performing comparator by more than 10% on eight datasets and by 23.2% on average (Fig. 4a). These results demonstrate the effectiveness of specialized PBT models across practical battery lifetime-prediction scenarios within pretraining-covered battery specifications. Notably, although the general PBT did not achieve the lowest MAPE on the three datasets (Fig. 3c), the specialized PBT achieved state-of-the-art performance after specialization. This result shows that the shared knowledge and parameter foundation encoded by the general PBT can be refined to capture target-specific degradation patterns even when the general model does not provide the best direct prediction.

Comparisons across the pretrained, specialized and from-scratch variants further show that target-dataset specialization does not benefit all model architectures equally. Although CPTransformer-T outperforms CPTransformer on 10 datasets and CPMLP-T outperforms CPMLP on 8 datasets, the pretrained models used without specialization and the variants trained from scratch using only target-dataset data achieve the second-best performance on 6 of the 12 datasets. This observation suggests that specialization can improve model performance but does not by itself ensure the strongest performance on individual target datasets. This conclusion is further supported by the results of BatLiNet on the 12 datasets (SI Table S3). BatLiNet does not match the specialized PBT across these datasets and produces a MAPE $\geq$ 50% for 6 datasets with limited data or aging conditions substantially different from those of LFP/graphite batteries, a trend also noted in the original BatLiNet study.[12] Because the original BatLiNet model was trained solely with information from LFP/graphite batteries, we further evaluate BatLiNet using the PBT pretraining data to isolate the effect of pretraining-data coverage. As summarized in SI Table S4, simply enlarging BatLiNet's pretraining set in this way improves its transfer performance for most datasets, but it still fails to match the performance of the specialized PBT, implying that integrating heterogeneous battery lifetime data is a non-trivial challenge. This outcome highlights both the difficulty and the necessity of developing an FM architecture that can effectively exploit diverse battery lifetime data, rather than merely relying on larger volumes of pretraining data.

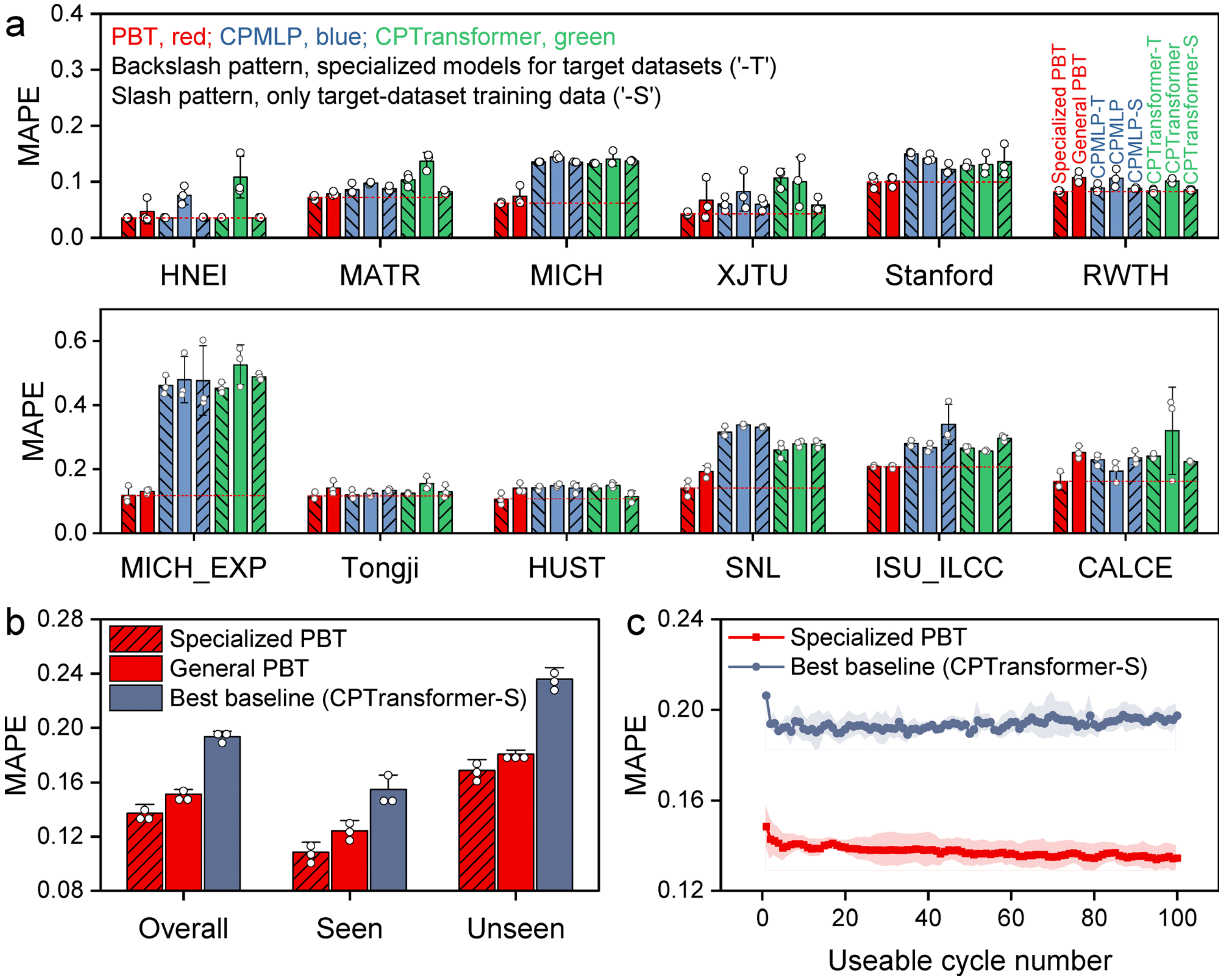

Fig. 4 Evaluation of specialized models for target datasets within pretraining-covered battery specifications. (a) MAPE of PBT, CPMLP and CPTransformer, together with their target-specialized variants (-T) and variants trained from scratch using only target-dataset training data (-S), across 12 target datasets. The suffix "-T" denotes a model specialized for an individual target dataset from its corresponding pretrained model, whereas "-S" denotes a model trained from scratch using only the corresponding target-dataset training data. Colors denote the model architecture (PBT, red; CPMLP, blue; CPTransformer, green). Solid bars denote pretrained models used without target-dataset specialization, bars with backslash hatching denote specialized models, and bars with slash hatching denote models trained from scratch. Dashed horizontal lines indicate the performance of specialized PBT as a visual reference. (b) MAPE of specialized PBT, the general PBT and the best-performing baseline (CPTransformer-S), aggregated across the 12 target datasets for all aging conditions (Overall) and for seen and unseen aging conditions. (c) MAPE of specialized PBT and the best-performing baseline (CPTransformer-S) as a function of the number of useable cycles. Bars in (a) and (b) and lines in (c) show means across three independent runs; individual points show the results of each run, and error bars and shaded regions indicate the corresponding standard deviations.

To assess performance across seen and unseen aging conditions, we aggregated the results across the 12 target datasets (Fig. 4b). Both the general PBT and the specialized PBT models outperformed the best-performing baseline. The general PBT reduced MAPE relative to this baseline by 21.9% overall and by 19.7% and 23.5% for seen and unseen aging conditions, respectively, indicating that it had acquired predictive knowledge relevant to these battery specifications during pretraining.

For each target dataset, the general PBT was then used to construct a specialized PBT that captures the degradation behavior specific to that dataset. Aggregated across the 12 target datasets, these specialized PBT models achieved an overall MAPE of 0.137. Relative to the general PBT, this corresponds to MAPE reductions of 9.3% overall and 12.8% and 6.6% for seen and unseen aging conditions, respectively. These gains demonstrate that the specialized PBT models capture degradation relationships specific to their respective target datasets while remaining applicable to aging conditions not represented during their construction. We further compared the specialized PBT models with the best-performing baseline across different numbers of useable cycles (Fig. 4c). The specialized PBT models maintained a lower MAPE than the best-performing baseline across all examined cycle numbers, demonstrating their consistent advantage across early-prediction settings with different amounts of cycling data.

In summary, these results establish the effectiveness of specialized PBT models for target datasets whose battery specifications are represented in the pretraining data. Constructed from the shared knowledge and parameter foundation provided by the general PBT, each specialized model captures degradation relationships specific to its target dataset and achieves consistent performance advantages across seen and unseen aging conditions and different amounts of early-cycle data. This capability supports practical applications such as rapid lifetime assessment in manufacturing and the optimization of fast-charging and formation protocols within covered battery specifications.

### Building specialized PBT models for target scenarios beyond pretraining-covered battery specifications

Having established the effectiveness of specialized PBT models for target datasets within pretraining-covered battery specifications, we next examine whether the same shared knowledge and parameter foundation can support specialized PBT models when the target battery specifications fall outside the pretraining coverage. We therefore construct and evaluate specialized PBT models for three target datasets comprising large-format industrial Li-ion, Na-ion and Zn-ion batteries. The Na-ion and Zn-ion datasets test the construction of specialized models for chemistries beyond LIBs, with their test sets consisting exclusively of unseen aging conditions. Specifically, the Na-ion test set contains operating conditions not represented in its training or validation data, whereas the Zn-ion test set differs from its training and validation data in physical format, electrolyte formulation and operating conditions. The industrial Li-ion test set comprises both seen and unseen aging conditions and represents two complementary industrial applications: predicting lifetime variations associated with manufacturing inconsistencies in established battery products and rapidly evaluating new operating conditions using early-cycle data, thereby reducing reliance on resource-intensive full-lifetime cycling tests. The numbers of test aging conditions and their classification as seen or unseen under different random seeds are summarized in SI Table S5. For controlled comparison, we apply the same specialization strategies to the pretrained CPMLP and CPTransformer models and include their variants trained from scratch using only the corresponding target-dataset training data. Direct use of the pretrained models without target-dataset specialization produces high prediction errors on all three datasets (SI Table S6), confirming substantial shifts from the pretraining distribution; these unsatisfactory direct-inference results are omitted from Fig. 5 for brevity. Together, the three datasets provide stringent tests of whether the PBT framework can support effective specialized models beyond its pretraining coverage.

Across the three target datasets beyond pretraining coverage, specialized PBT achieved the lowest mean MAPE, reducing MAPE relative to the best-performing comparator by 42.1%, 19.0%, and 31.3% for the industrial Li-ion, Na-ion, and Zn-ion battery datasets, respectively (Fig. 5a–c). When the results were stratified by aging-condition familiarity, specialized PBT also achieved the lowest mean MAPE under both seen and unseen conditions (SI Table S7), supporting its adaptability to non-Li-ion chemistries and industrial development of Li-ion cells. The performance trends across varying numbers of useable cycles (Fig. 5d–f) further show that specialized PBT yields the best results in nearly all early prediction settings, indicating that specialized PBT models maintain strong performance across different amounts of early-cycle information. Specialized PBT also outperforms BatLiNet (SI Tables S3 and S4). In addition, CPMLP-T and CPTransformer-T may perform worse than models trained solely on the target datasets, suggesting that the gains of specialized PBT stem from its ability to adapt broadly learned battery lifetime knowledge to target scenarios.

Collectively, these findings establish that PBT supports the construction of effective specialized models even when the target battery specifications fall outside the pretraining coverage. This capability extends the applicability of PBT from battery systems represented in the pretraining data to new chemistries and industrial battery designs, providing a basis for accelerating the development and deployment of next-generation energy storage materials and systems.

### Origins of PBT performance

Theoretical analyses of transfer learning[40,41] indicate that the target-domain performance of specialized models is constrained by two interrelated factors: the predictive capability of the pretrained model and the extent to which the pretraining distributions cover degradation patterns relevant to the target domain. Guided by this framework, we investigated the origins of PBT performance from both perspectives. To examine the first factor, we evaluated how PBT's architecture and aging-condition inputs affect its performance on held-out pretraining data through a series of ablation studies (Fig. 6a). Specifically, we removed either the hard encoder (PBT w/o HE) or the soft encoder (PBT w/o SE). We also constructed PBT w/o LLM by replacing the LLM-based aging network with learned lookup embeddings covering all aging factors amenable to this representation; implementation details of this lookup-based approach are provided in SI Note S5. To assess the contribution of different types of aging-condition information, we separately retained only battery specifications or only formation and operating conditions in the text-based prompt. We further included three architectural baselines. CPTransformer removes BatteryMoE from PBT, whereas DeepSeekMoE replaces BatteryMoE with a generic MoE architecture.[50] CPTransformer-I augments CPTransformer with the same LLM-generated aging-condition embeddings used by PBT (see SI Note S6 for implementation details), allowing us to distinguish the contribution of aging-condition information from the way in which BatteryMoE uses this information. To examine the second factor, we varied the amount of pretraining data from 25% to 100% while keeping the test sets fixed (Fig. 6b). This design isolates the effect of pretraining-data scale on performance over held-out pretraining data. Improved performance with increasing data volume would indicate more effective consolidation of degradation patterns across the pretraining datasets and may reflect broader coverage of patterns relevant to downstream target datasets.

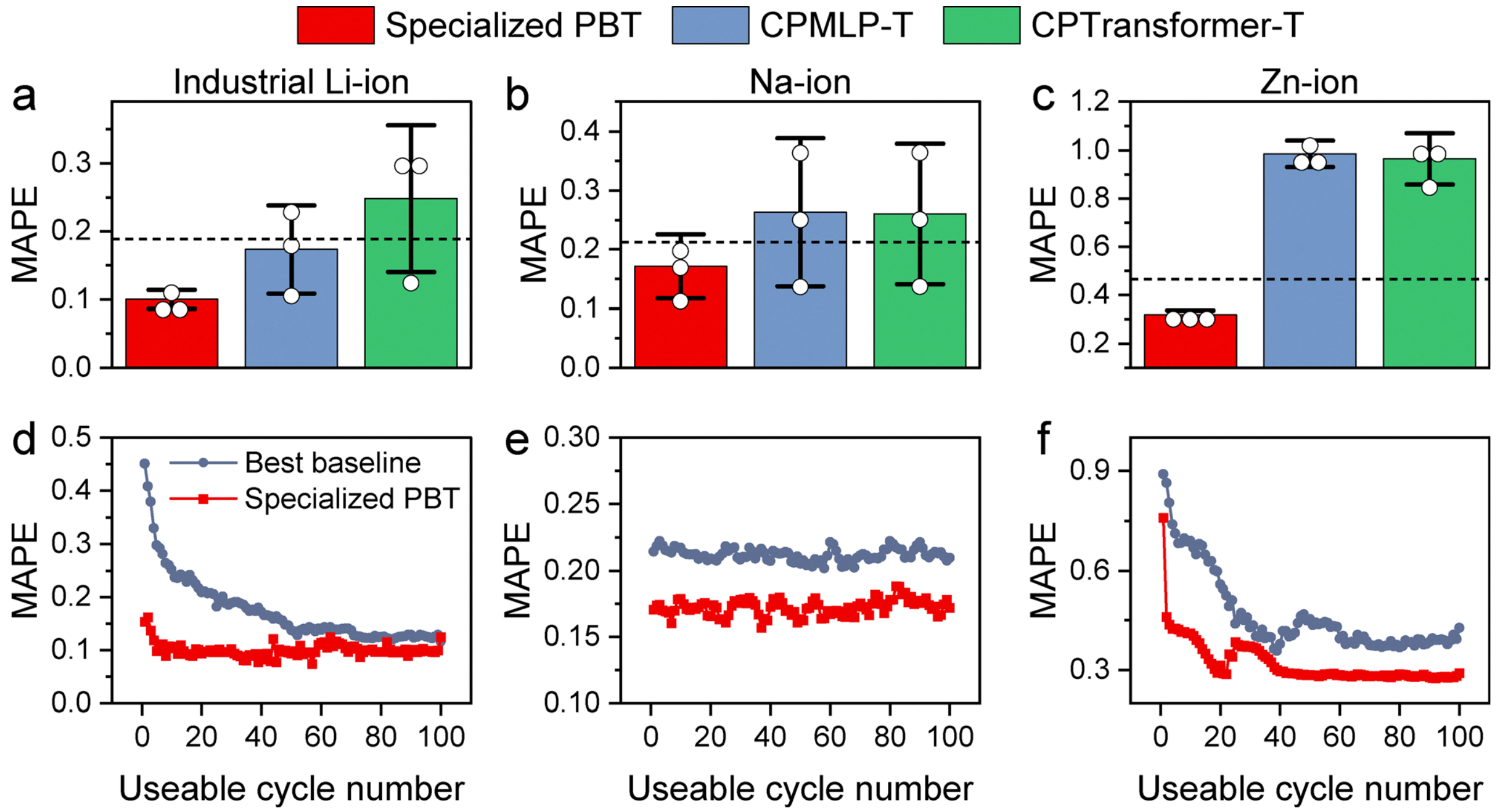

Fig. 5 Evaluation of specialized models for target datasets beyond pretraining-covered battery specifications. (a)–(c) MAPE of the specialized PBT, CPMLP-T and CPTransformer-T on the industrial Li-ion (a), Na-ion (b) and Zn-ion (c) battery datasets. CPMLP-T and CPTransformer-T denote models specialized for the corresponding target dataset from their respective pretrained models. The dashed horizontal line in each panel indicates the best result obtained by CPMLP-S or CPTransformer-S, where the suffix "-S" denotes a model trained from scratch using only the corresponding target-dataset training data. Bars show means and error bars indicate standard deviations across three independent runs. (d)–(f), Model performance as a function of the number of useable cycles for the industrial Li-ion (d), Na-ion (e) and Zn-ion (f) battery datasets. Because 17 of the 27 batteries in the industrial Li-ion dataset lack the voltage–current data at the 100th cycle, the number of evaluable batteries decreases abruptly at this point, which may produce a discontinuity in MAPE.

The ablation results show that both knowledge-encoding pathways contribute to PBT's performance on held-out pretraining data (Fig. 6a). Removing either the soft encoder or the hard encoder increased MAPE relative to the full PBT, indicating that the two encoders play distinct but complementary roles. The soft encoder converts descriptions of aging conditions into contextual representations that guide expert weighting, whereas the hard encoder uses established battery attributes to retain experts consistent with these attributes. Their combined use allows PBT to incorporate aging-condition information while keeping expert routing consistent with battery knowledge. We also examined the importance of the LLM-based representation by replacing it with learned lookup embeddings. This replacement increased MAPE by 13.9%, indicating that the full text-based representation pipeline produces more informative aging-condition representations than the lookup scheme tested here. The baseline comparisons further clarify the respective contributions of the model architecture and aging-condition information. PBT outperformed CPTransformer, which does not contain BatteryMoE. DeepSeekMoE also failed to match PBT, showing that introducing a generic MoE architecture alone is insufficient to achieve the same performance. When CPTransformer was supplied with the same LLM-generated aging-condition embeddings as PBT, the resulting CPTransformer-I model performed better than CPTransformer but remained less accurate than PBT. Together, these comparisons show that informative aging-condition representations improve prediction, while BatteryMoE provides an effective mechanism for translating this information into battery-knowledge-guided expert routing.

After examining how PBT incorporates aging-condition information into expert routing, we next assessed the contribution of each information category to performance on held-out pretraining data. Retaining only battery specifications or only formation and operating conditions in the prompt increased MAPE relative to using the complete set of aging-condition information (Fig. 6a), indicating that the two categories provide complementary information. Battery specifications characterize the material and design attributes of a cell, whereas formation and operating conditions describe the protocols and external conditions applied during formation and subsequent cycling. These results show that PBT benefits from jointly incorporating battery specifications and formation as well as operating conditions, rather than relying on either information category alone.

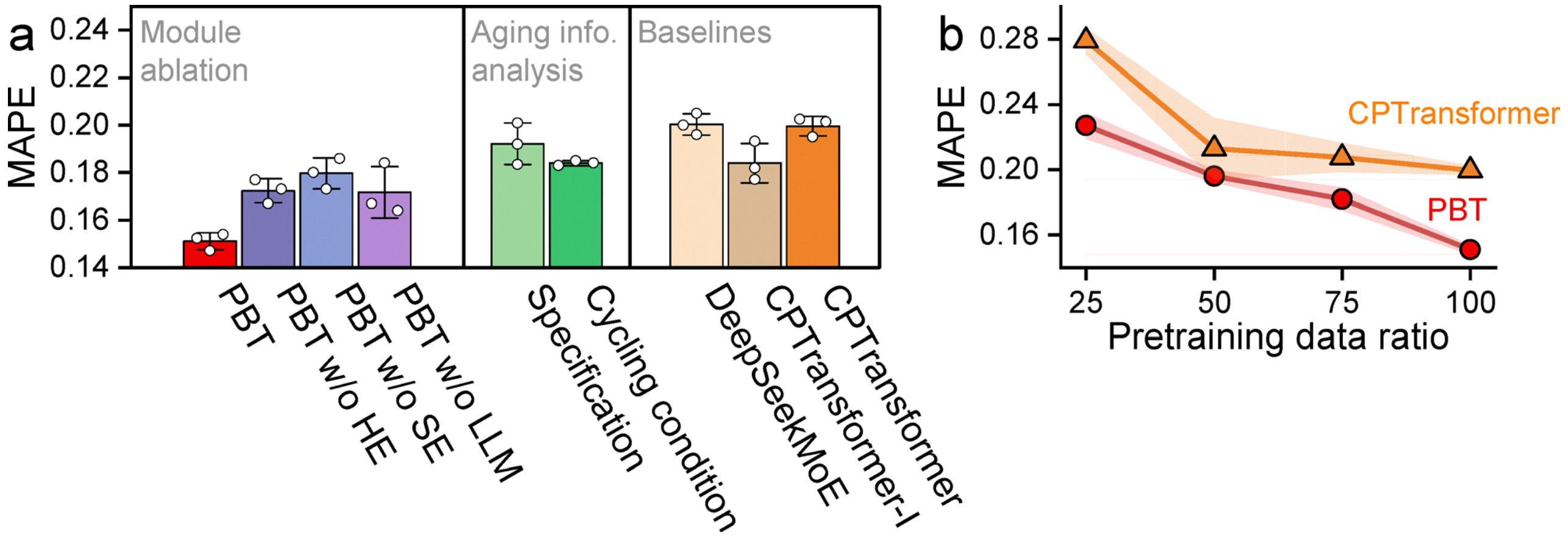

Fig. 6 Ablation and data-scaling studies. (a) Ablation of the principal components of PBT and the aging-condition information provided in the prompt, compared with the CPTransformer, CPTransformer-I and DeepSeekMoE baselines. "PBT w/o HE" and "PBT w/o SE" denote variants without the hard encoder and soft encoder, respectively. In "PBT w/o LLM", the LLM-based aging embedder is replaced by learned lookup embeddings representing the cathode chemistry, anode chemistry, operating temperature and physical format. The "Specification" and "Cycling condition" variants use prompts containing only battery specifications and only formation and operating conditions, respectively. CPTransformer removes BatteryMoE from PBT, CPTransformer-I augments CPTransformer with the same LLM-generated aging-condition embeddings used by PBT, and DeepSeekMoE replaces BatteryMoE with a generic MoE architecture. (b) Performance of PBT and CPTransformer when pretrained using 25%, 50%, 75% and 100% of the available pretraining data. Bars in (a) and lines in (b) show means across three independent runs; individual points in (a) show the result of each run, and error bars in (a) and shaded regions in (b) indicate the corresponding standard deviations.

Finally, we examined how pretraining-data scale affects performance on the fixed held-out pretraining test sets (Fig. 6b). PBT achieved a lower MAPE than CPTransformer at all four data ratios. Moreover, PBT pretrained with 50% and 75% of the available data outperformed CPTransformer pretrained with 75% and 100%, respectively. Because CPTransformer is obtained by removing BatteryMoE from PBT, these comparisons indicate that BatteryMoE enables PBT to use the available pretraining data more efficiently. Within PBT, MAPE also decreased consistently as the volume of pretraining data increased. With the same held-out test sets used at every data ratio, this trend indicates that additional pretraining data improved the learning of cycling-pattern–lifetime relationships across the heterogeneous datasets. By broadening the coverage of degradation patterns, a larger pretraining corpus may also increase the likelihood that PBT acquires knowledge relevant to subsequent target datasets. Together, these results identify both the data efficiency enabled by BatteryMoE and the broader degradation-pattern coverage provided by increasing pretraining-data scale as important contributors to PBT's performance.

Together, these findings support a statistical rationale for PBT's performance based on the joint contributions of structured knowledge organization, aging-condition information and pretraining-data coverage. First, BatteryMoE provides a mechanism for encoding heterogeneous cycling-pattern–lifetime relationships into model parameters through battery-knowledge-guided expert routing. Second, informative aging-condition representations provide the contextual information needed for aging-condition-dependent expert selection and weighting based on battery specifications, formation protocols and operating conditions. Third, a larger volume of pretraining data broadens the statistical coverage of degradation patterns, increasing the likelihood that knowledge relevant to subsequent target datasets is represented in the general PBT.

## Discussion and conclusions

This work introduces the PBT, to our knowledge the first foundation model for battery cycle-life prediction under data scarcity and heterogeneity. PBT is an integrated framework comprising a general PBT and specialized PBT models. At its core, BatteryMoE enables the general PBT to consolidate battery lifetime knowledge from 13 heterogeneous lithium-ion battery datasets through battery-knowledge-guided expert routing. Specialized PBT models are then constructed from this shared knowledge and parameter foundation using limited data from individual target scenarios. Across 15 downstream datasets spanning diverse charge–discharge protocols, temperatures, stress factors, formation procedures, industrial battery designs and non-lithium-ion chemistries, the specialized PBT models consistently outperform competing approaches. Neural baselines pretrained on the same data and specialized using the same procedures do not achieve comparably consistent gains, indicating that effective specialization depends on how heterogeneous lifetime knowledge is organized rather than on pretraining-data scale and diversity alone. The ablation and data-scaling results further associate PBT's performance with complementary aging-condition information, knowledge-guided expert routing and broader coverage of degradation patterns in the pretraining data. Together, these findings establish PBT as a practical framework in which the general model provides reusable battery lifetime knowledge and the specialized models capture target-specific degradation behavior.

From a practical research and development perspective, PBT could support both the development of next-generation batteries and manufacturing quality control for commercial cells. For a target application, the general PBT would first be specialized using a limited set of cells with full-life labels and then integrated into the corresponding decision workflow. In battery development, for example, the specialized PBT could be embedded within a closed-loop optimization framework,[15] in which lifetime predictions obtained from early-cycle voltage–current data are used to deprioritize less promising candidates and guide the selection of subsequent experiments. This approach could accelerate the screening of candidate material compositions,[82] formation protocols[26] and operating protocols.[15,16] Alternatively, PBT-generated lifetime predictions could provide pseudo-labels for training a recipe-based surrogate model that maps design variables directly to lifetime, thereby enabling exploration of a broader candidate space for experimental validation.[33] Full-life measurements from these selected candidates could subsequently be fed back to update the specialized PBT and guide the next optimization cycle. In manufacturing, because full-life cycling is impractical for every cell, a product- or production-line-specific PBT could use routinely collected early-cycle test data to flag cells with low predicted lifetime or batches exhibiting anomalous lifetime distributions for targeted follow-up testing before lot release. By directing full-life characterization and confirmatory testing towards the most informative or potentially non-conforming candidates, PBT could improve both the efficiency of battery development and the effectiveness of manufacturing quality assurance.

The cross-domain evaluations also delineate PBT's domain of applicability and reveal two principal limitations. First, predictive accuracy is bounded by the information contained in the model inputs. Prediction errors can therefore remain high when only one cycle is available, as observed in Fig. 5f. Conversely, when many cycles are available, the performance gap between specialized PBT and strong baselines may narrow because the measurements provide sufficient predictive information for multiple models. For example, the best-performing baseline approaches specialized PBT when nearly 100 cycles are used for prediction in the industrial Li-ion dataset (Fig. 5d). Second, the knowledge available for target-scenario specialization is constrained by the degradation patterns represented in the training data, consistent with transfer-learning theory.[40,41] The general PBT outperforms the best-performing baseline by 23.5% for unseen aging conditions aggregated across the 12 downstream Li-ion datasets whose battery specifications are represented in pretraining (Fig. 4b), but its direct application produces high prediction errors under the substantial specification shifts represented by the industrial Li-ion, Na-ion and Zn-ion datasets (SI Table S6). After specialization with limited labelled target-domain data, however, the corresponding specialized PBT models outperform the strongest comparators by 42.1%, 19.0% and 31.3% on these three datasets, respectively (Fig. 5a–c). The PBT framework can therefore support the construction of specialized models for target scenarios both within and beyond pretraining-covered battery specifications, but the attainable performance of these models remains limited by the predictive content of the input measurements and by how well the available training data cover degradation patterns relevant to the target domain.

Several directions deserve further exploration to expand PBT's capabilities and improve the reliability of its predictions. One promising direction is to incorporate more comprehensive descriptions of aging conditions. The ablation results show that battery specifications and formation and operating conditions provide complementary information (Fig. 6a), suggesting that consistently recorded manufacturing variables, such as coating-process parameters and stack pressure, could provide additional cues for expert routing and prediction. Improving predictive uncertainty estimation represents another promising direction for practical deployment. Monte Carlo dropout[83] generally assigns higher uncertainty to less accurate predictions (SI Note S7 and Fig. S4), providing preliminary evidence that the estimated uncertainty can help identify potentially unreliable outputs. Further work could improve uncertainty calibration and evaluate it systematically across aging conditions. Meanwhile, PBT is developed on voltage-current data. Incorporating complementary measurements beyond voltage–current profiles may provide a promising route to further improving early-cycle battery lifetime prediction because this can offer information beyond the voltage–current data.

Beyond its specific implementation in PBT, the two-level knowledge injection strategy in BatteryMoE, which combines a representation-guiding soft encoder with a constraint-imposing hard encoder, may provide a useful design pattern for other prediction problems characterized by data scarcity and heterogeneity, such as battery capacity estimation, solar cell lifetime prediction and wind power forecasting. In such settings, the soft encoder could incorporate domain-relevant factors, such as material composition, processing history and geographic position, whereas the hard encoder could introduce physically meaningful routing constraints derived from established governing factors. This perspective suggests that BatteryMoE is not limited to battery life prediction, but may offer a generalizable architecture for embedding domain knowledge into foundation models for a broader class of sustainable energy applications.

Overall, this study shows that tailored foundation models can provide an effective route to addressing the long-standing challenges of data scarcity and heterogeneity in battery life prediction. More broadly, it suggests that scientific machine learning for energy systems should move towards architectures that can structure fragmented experimental observations into transferable, specialization-ready representations. In this sense, battery life prediction provides a representative example of how domain knowledge can guide foundation-model design and make scarce and heterogeneous experimental data more reusable across sustainable energy applications.

## Methods

The PBT is implemented as a battery-knowledge-encoded neural architecture with BatteryMoE as its core module. As illustrated in Fig. 2b, PBT consists of a BatteryMoE-CyclePatch layer, a BatteryMoE intra-cycle encoder with $L_1$ layers, a BatteryMoE-inter-cycle encoder with $L_2$ layers, and a projection head. We first describe BatteryMoE, followed by the individual model components.

**BatteryMoE.** BatteryMoE is a mixture-of-experts layer that incorporates battery knowledge through a soft encoder and a hard encoder. The soft encoder embeds predefined aging factors that dominate cycling data patterns into vectors, whereas the hard encoder imposes routing constraints based on well-established categorical or continuous aging factors. These factors—spanning battery specifications, formation and operation—are embedded into vectors as follows:

$$e = E_{\mathrm{ae}}\,(\text{specification, formation, operation}), \quad (1)$$

where $E_{\mathrm{ae}}$ denotes an aging embedder and $e$ is the resulting vector representation. To construct a universal embedder applicable to arbitrary combinations of aging factors, we devise a large language model (LLM)-based aging network that produces token embeddings from a textual prompt describing the factors:

$$\bar{E} = \mathrm{LLM}\,(\text{prompt}), \quad (2)$$

$$e = \mathrm{lastValid}\,(\bar{E}) \quad (3)$$

where prompt (SI Fig. S1) is a text description of the aging factors, $\bar{E} \in \mathbb{R}^{L\times d_{\mathrm{LLM}}}$ is the embedding of tokens obtained from the prompt, and $e \in \mathbb{R}^{d_{\mathrm{LLM}}}$ is the embedding of the last valid token in $\bar{E}$. The embedding of the final valid token $e$ is used as the aging-condition representation because, under causal attention, this token can attend to the preceding tokens in the prompt. In the present study, Llama-3.1-8B-Instruct[84] is used as the LLM to embed the prompt. The LLM parameters are frozen during both PBT pretraining and transfer learning to mitigate overfitting and training instability under limited-data regimes. Notably, because the aging conditions are typically predetermined, the corresponding embedding $e$ can be generated before the degradation test and cached for subsequent use, thereby avoiding repeated LLM inference during model training and deployment. A nonlinear transformation is then applied to obtain informative representations:

$$\hat{e} = \mathrm{LeakyReLU}(We + b), \quad (4)$$

where $W \in \mathbb{R}^{d_{ff}\times d_{\mathrm{LLM}}}$ and $b \in \mathbb{R}^{d_{\mathrm{ff}}}$. The processed embedding $\hat{e}$ is shared across BatteryMoE layers and used in a distinct linear layer of each BatteryMoE layer to determine expert weights as follows:

$$g = [g_1, g_2, g_3, \ldots, g_{K_s}] = W_{l,2}\hat{e} + b_{l,2}, \quad (5)$$

where $g \in \mathbb{R}^{K_s}$ denotes the initial unnormalized expert weights for $K_s$ expert networks, $W_{l,2} \in \mathbb{R}^{K_s\times d_{ff}}$ and $b_{l,2} \in \mathbb{R}^{K_s}$ are learnable parameters for the $l$th BatteryMoE layer and $K_s$ is the number of specialized expert networks in a BatteryMoE layer.

The hard encoder (blue in Fig. 2a) imposes knowledge-based routing constraints within the MoE layer. We apply hard routing to four aging factors with well-defined categories or physically interpretable ranges across the collected datasets: cathode, anode, physical format and operating temperature. Other factors, such as formation and charge protocols, are not hard-coded because their categorical definitions are less clear across studies. Under this scheme, only the expert weights associated with the matching categories or temperature range are retained for modeling a given battery:

$$\tilde{g} = \sigma(g), \quad (6)$$

$$\bar{g} = \left[\bar{g}_1, \bar{g}_2, \bar{g}_3, \ldots, \bar{g}_{K_s}\right] = \frac{\tilde{g}}{\sum_{i=1}^{K_s} \tilde{g}_i}, \quad (7)$$

where $\sigma(\cdot)$ denotes the knowledge-based weight selection mechanism used to set irrelevant expert weights to zero, and $\bar{g}$ is the resulting normalized expert-weight vector used in BatteryMoE to aggregate outputs of expert networks. Specifically, for categorical factors (cathode, anode, and physical format), only expert networks corresponding to the selected category are used. For temperature, which is treated as a continuous variable, experts whose specialized temperatures fall within $\pm 5\ ^{\circ}\mathrm{C}$ of the input are selected. The resulting knowledge-informed expert weight $\bar{g}$ is the output of the battery gate (Fig. 2a) and is subsequently used to aggregate outputs from expert networks:

$$M^l = \sum_{i=1}^{K_g} F_i\left(M^{l-1}\right) + \sum_{j=1}^{K_s} \bar{g}_j F_{j+K_g}\left(M^{l-1}\right), \quad (8)$$

where $K_g$ denotes the number of general experts and $F_i(\cdot)$ is an expert network. This formulation makes BatteryMoE a plug-and-play component compatible with various expert architectures $F(\cdot)$. For brevity, in the following subsections, we refer to a BatteryMoE layer as BatteryMoE($F(\cdot)$) when $F(\cdot)$ is used as the expert network architecture. To ensure effective expert specialization, the number of experts for each category is determined by rounding the number of training batteries to the nearest hundred and assigning one expert per 100 batteries, with a minimum of one expert (*e.g.*, 151 → two experts, 101 → still one expert). This adaptive allocation scheme supports fine-scale specialization through data-driven learning for categories with relatively abundant training data.

**BatteryMoE-CyclePatch.** Motivated by the previous success of CyclePatch,[13] BatteryMoE-CyclePatch is employed to patch the cycling data for each cycle into a token. Let $X_{1:S} \in \mathbb{R}^{(300\times S)\times 3}$ denote the voltage, current and capacity sequence of the first $S$ cycles (each cycle is resampled to 300 data points; see SI Note

S8). The process of BatteryMoE-CyclePatch can be described as follows:

$$[X_1, X_2, X_3, \ldots, X_S] = \mathrm{Segment}(X_{1:S}), \quad (9)$$

$$\hat{X}_i = \mathrm{BatteryMoE}(\mathrm{Linear}(\mathrm{Flatten}(X_i))), \quad (10)$$

where $X_i \in \mathbb{R}^{300\times 3}$, $\mathrm{Flatten}(X_i) \in \mathbb{R}^{900}$, and $\mathrm{Linear}(\cdot)$ is a linear layer that projects $\mathrm{Flatten}(X_i)$ to $\hat{X}_i \in \mathbb{R}^d$. In this context, the recurring cycling data patterns are explicitly segmented into distinct tokens $\hat{X} = [\hat{X}_1, \hat{X}_2, \ldots, \hat{X}_S]$.

**BatteryMoE-Intra-cycle encoder.** Following CyclePatch, an intra-cycle encoder is employed to mine useful high-level representations for each cycle. This process for one intra-cycle encoder layer can be formulated as follows:

$$H_i^l = \mathrm{LN}(\mathrm{BatteryMoE}(\mathrm{FFN}(H_i^{l-1})) + H_i^{l-1}), \quad (11)$$

where $\mathrm{LN}(\cdot)$ denotes layer normalization,[85] $\mathrm{FFN}(\cdot)$ is a feedforward network with a GELU[86] activation function, $H_i^l \in \mathbb{R}^d$, and $H_i^0 = \hat{X}_i$. After stacking $L_1$ intra-cycle encoder layers, we obtain the cycle-token embeddings for the first $S$ cycles, denoted by $H^{L_1}$. These embeddings are then fed into the inter-cycle encoder to capture interactions across cycles. For efficient batch operation, all $H^{L_1}$ vectors are zero-padded to a length of 100, corresponding to the maximum number of usable input cycles, resulting in $H^{L_1} \in \mathbb{R}^{100\times d}$.

**BatteryMoE-Inter-cycle encoder.** The inter-cycle encoder adopts a transformer[87] encoder framework for modeling information across cycle tokens. First, positional encoding is used to incorporate position information into the intra-cycle encoder output:

$$\bar{H}^{L_1} = \mathrm{PE}(H^{L_1}) + H^{L_1}, \quad (12)$$

where $\mathrm{PE}(\cdot)$ is the positional encoding introduced in the original transformer paper.[87] Then, $\bar{H}^{L_1}$ is input into the inter-cycle encoder, where the process for one layer is as follows:

$$U^l = \mathrm{LN}(\mathrm{Attention}(P^{l-1}, P^{l-1}, P^{l-1}) + P^{l-1}), \quad (13)$$

$$P^l = \mathrm{LN}(\mathrm{BatteryMoE}(\mathrm{FFN}(U^l)) + U^l), \quad (14)$$

where $\mathrm{Attention}(\cdot,\cdot,\cdot)$ is the self-attention[87] operation, $P^0 = \bar{H}^{L_1}$, and $P^l \in \mathbb{R}^{100\times d}$ is the output of the $l$th inter-cycle encoder layer. Note that the attention mask is adopted to ignore the padded tokens in $\mathrm{Attention}(\cdot,\cdot,\cdot)$.

**Projection head.** After the intra-cycle and inter-cycle encoders, the resulting sequence representation $P^{L_2}$ contains both intra-cycle and inter-cycle information. The last nonmasked vector in $P^{L_2}$ is denoted by $z$ and is used as the input to the projection head. The final prediction $\hat{y}$ is produced by a projection head comprising five general linear layers and five specialized linear layers:

$$[\hat{g}_1, \hat{g}_2, \hat{g}_3, \hat{g}_4, \hat{g}_5] = G(z), \quad (15)$$

$$\hat{y} = \sum_{i=1}^{5} F_i(z) + \sum_{j=1}^{5} \hat{g}_j F_{j+5}(z), \quad (16)$$

**Model optimization.** The PBT is optimized by minimizing the mean squared error $\mathscr{L} = (\hat{y} - y)^2$ using the AdamW[88] optimizer. The hyperparameters that yield the best results for the validation set are selected. The search ranges of the hyperparameters and selected hyperparameters for pretraining the PBT are listed in SI Table S8. The hardware configuration used for model training and evaluation is summarized in SI Table S9. The computational characteristics of PBT, including the number of trainable parameters, total training time, and per-sample serial inference latency, are reported in SI Table S10.

**PBT specialization.** After pretraining, the resulting model serves as a general PBT that encodes battery lifetime knowledge learned from heterogeneous pretraining datasets. For each target dataset, the general PBT is specialized using the available target-scenario training data. Two transfer learning strategies are considered: fine-tuning, where pretrained model parameters are updated, and adapter tuning, where lightweight trainable modules are inserted while the pretrained backbone is kept fixed. For fair comparison, the same specialization strategies are applied to CPMLP and CPTransformer. For each model and target dataset, the strategy that achieves the best validation performance is used for testing. Additional implementation details are provided in SI Note S3.

**Evaluation metric.** The MAPE is employed to evaluate model performance and is computed as follows:

$$\mathrm{MAPE} = \frac{1}{S_{\mathrm{test}}} \sum_{i=1}^{S_{\mathrm{test}}} \frac{1}{N_i} \sum_{j=1}^{N_i} \frac{|y_j - \hat{y}_j|}{y_j}, \quad (17)$$

where $y_j$ and $\hat{y}_j$ denote the ground-truth and predicted battery lifetimes of one sample of the $j$th test cell, respectively, $S_{\mathrm{test}}$ denotes the number of distinct aging conditions in the test set, and $N_i$ means the number of cells in the $i$th aging condition.

## Author contributions

T.-Y. Zhang, J. Huang, R. Tan and J. Li conceived the study. R. Tan designed the model and conducted the AI analysis with the help of X. Hong and J. Li. R. Tan, J. Huang and T.-Y. Zhang wrote the paper, with contributions from all the authors. T.-Y. Zhang and J. Huang managed the project and guided the research.

## Conflicts of interest

The authors declare a competing interest: a Chinese patent application related to the method reported in this work has been filed under application number 202610578824X.

## Data availability

All the datasets used in the present study are publicly available, including CALCE,[66,70] MATR,[15,16] RWTH,[68] HUST,[21] Tongji,[64] Stanford,[60] XJTU,[65] and ISU_ILCC.[57] HNEI,[62] SNL,[59] MICH_EXP,[61] MICH,[26] and UL-PUR[58,67] are hosted in the BatteryArchive (https://www.batteryarchive.org/). The Zn-ion, Na-ion and industrial LIB datasets are available through the BatteryLife repository[13] (https://github.com/Ruifeng-Tan/BatteryLife).

Code availability: the code for the modeling work and data preprocessing is available at https://github.com/Ruifeng-Tan/PBT. The weights of the general PBT are available at https://zenodo.org/records/17972780.

Supplementary information (SI), containing additional methodological details and supporting results. See DOI: https://doi.org/10.1039/d6ee03446b.

## Acknowledgements

The authors acknowledge financial support from the National Key R&D Program of China (No. 2023YFB2503600). This work is also supported by research grants from the National Natural Science Foundation of China (No. 92372109 and 52207230) and the Guangdong Provincial Talent Program (No. 2024TQ08X366). We also acknowledge support from the Wilson Tang Brilliant Energy Science and Technology Lab (BEST Lab) at Hong Kong University of Science and Technology (Guangzhou).